\documentclass{article}
\usepackage{graphicx} 
\usepackage{geometry}
\usepackage[ruled, vlined, linesnumbered]{algorithm2e}
\usepackage{listings}
\usepackage{xcolor}
\usepackage{microtype}
\usepackage{float}
\usepackage{xurl} 
\usepackage[colorlinks=true, urlcolor=blue]{hyperref} 

\colorlet{punct}{red!60!black}
\definecolor{background}{HTML}{EEEEEE}
\definecolor{delim}{RGB}{20,105,176}
\colorlet{numb}{magenta!60!black}

\lstdefinelanguage{json}{
    basicstyle=\normalfont\ttfamily,
    numbers=left,
    numberstyle=\scriptsize,
    stepnumber=1,
    numbersep=8pt,
    showstringspaces=false,
    breaklines=true,
    frame=lines,
    backgroundcolor=\color{background},
    literate=
     *{0}{{{\color{numb}0}}}{1}
      {1}{{{\color{numb}1}}}{1}
      {2}{{{\color{numb}2}}}{1}
      {3}{{{\color{numb}3}}}{1}
      {4}{{{\color{numb}4}}}{1}
      {5}{{{\color{numb}5}}}{1}
      {6}{{{\color{numb}6}}}{1}
      {7}{{{\color{numb}7}}}{1}
      {8}{{{\color{numb}8}}}{1}
      {9}{{{\color{numb}9}}}{1}
      {:}{{{\color{punct}{:}}}}{1}
      {,}{{{\color{punct}{,}}}}{1}
      {\{}{{{\color{delim}{\{}}}}{1}
      {\}}{{{\color{delim}{\}}}}}{1}
      {[}{{{\color{delim}{[}}}}{1}
      {]}{{{\color{delim}{]}}}}{1},
}

\title{A Hierarchical Tree-based approach for creating Configurable and Static Deep Research Agent (Static-DRA)}
\author{
  Saurav Prateek \\
}
\date{December 2025}

\begin{document}

\maketitle

\begin{abstract}
The advancement in Large Language Models has driven the creation of complex agentic systems, such as Deep Research Agents (DRAs), to overcome the limitations of static Retrieval Augmented Generation (RAG) pipelines in handling complex, multi-turn research tasks. This paper introduces the Static Deep Research Agent (Static-DRA), a novel solution built upon a configurable and hierarchical Tree-based static workflow.

The core contribution is the integration of two user-tunable parameters, Depth and Breadth, which provide granular control over the research intensity. This design allows end-users to consciously balance the desired quality and comprehensiveness of the research report against the associated computational cost of Large Language Model (LLM) interactions. The agent's architecture, comprising Supervisor, Independent, and Worker agents, facilitates effective multi-hop information retrieval and parallel sub-topic investigation.

We evaluate the Static-DRA against the established DeepResearch Bench using the RACE (Reference-based Adaptive Criteria-driven Evaluation) framework. Configured with a \textbf{depth} of \textbf{2} and a \textbf{breadth} of \textbf{5}, and powered by the \textbf{gemini-2.5-pro} model, the agent achieved an overall score of \textbf{34.72}. Our experiments validate that increasing the configured Depth and Breadth parameters results in a more in-depth research process and a correspondingly higher evaluation score. The Static-DRA offers a pragmatic and resource-aware solution, empowering users with transparent control over the deep research process. The entire source code, outputs and benchmark results are open-sourced at \url{https://github.com/SauravP97/Static-Deep-Research/}

\end{abstract}

\section{Introduction}

Advancement  in the Large Language Models has led to the development of multiple complex agentic systems that can research on complex topics and perform reasoning tasks. The conceptual lineage of Deep Research Agent begins with the limitations of early RAG (Retrieval Augmented Generation) \cite{lewis2021retrievalaugmentedgenerationknowledgeintensivenlp} architectures. These systems were characterized by a static, two-stage pipeline: a retriever would fetch a set of relevant documents, and a generator (a Large Language Model) would produce an answer based solely on those retrieved passages. This static approach is fundamentally limited when faced with complex, multi-faceted research queries that require iterative refinement, information synthesis from disparate sources, or interaction with the environment \cite{huang2025deepresearchagentssystematic}. There has been continuous research in improving these systems through advanced RAG techniques like Corrective-RAG \cite{yan2024correctiveretrievalaugmentedgeneration}, Self-RAG \cite{selfretrievalaugmentedgeneration}, FLARE \cite{zhang2024enhancinglargelanguagemodel}, IAG \cite{zhang2023iaginductionaugmentedgenerationframework}, SAIL \cite{luo2023sailsearchaugmentedinstructionlearning} and many more.

To overcome the above mentioned limitations of the agentic system, Deep Research Agents were introduced which are designed to tackle complex, multi-turn informational research tasks by leveraging a combination of dynamic reasoning, adaptive long-horizon planning, multi-hop information retrieval, iterative tool use, and the generation of structured analytical reports. The paper \cite{zhang2025deepresearchsurveyautonomous} complements this by defining the operational pipeline of a deep research agent. It frames them as a solution to the "internal knowledge boundaries" of LLMs and outlines a standard, four-stage workflow: Planning, Question Developing (Decomposing the main query), Web Exploration (Iterative retrieval and tool use) and Report Generation (Synthesis). Many organizations like OpenAI \cite{openaideepresearch}, Google’s Gemini \cite{geminideepresearch}, Grok \cite{xaideepresearch}, and Perplexity \cite{perplexitydeepresearch}, LangChain \cite{langchaindeepresearch} have developed their own deep research agents which perform in-depth research.

The paper \cite{huang2025deepresearchagentssystematic} states a clear difference between the \textbf{Static} and \textbf{Dynamic} deep research agents.

\begin{enumerate}
    \item \textbf{Static Workflows}: Static workflows rely on manually predefined task pipelines, decomposing research processes into sequential subtasks executed by dedicated agents.
    \item \textbf{Dynamic Workflows}: Dynamic workflows support adaptive task planning, allowing agents to dynamically reconfigure task structures based on iterative feedback and evolving contexts.
\end{enumerate}

We introduce a \textbf{configurable} and \textbf{hierarchical Tree-based} static Deep Research Agent which has the capability to perform deep research on the topic provided by the end user. The agent is built on a static workflow which can be configured by the end user to decide how deep they want the agent to research a particular topic. We introduce two configurable parameters which can be tuned by the end-user to decide on the depth of the research to be done by our Deep Research Agent.

We aim to build a configurable agent with deep research capability which can be controlled by the end user. Integrating with a Large Language Model (LLM) costs money and the pricing is calculated on multiple parameters (input/output tokens, number of requests etc.). The Deep Research Agent described in this paper allows users to control the number of requests they make to an integrated LLM at their will. If a research topic needs an in-depth research the breadth and depth value can be configured to higher value for the agent to perform a detailed research. While for simpler topics which require a shallow research, the depth and breadth can be configured to a relatively lower value for the agent to perform a shallow research and hence having a decent control on the LLM pricing.

The agent framework has a \textbf{supervisor} agent which can further spawn \textbf{independent agents} to research on the sub-topic independently. If a research topic can not be further split into sub-research topics then a \textbf{worker} agent goes ahead and performs a research on that sub-topic. We also use a \textbf{Web Search Tool} to look for citations / references and summarized search results for research sub-topics. The research output of the spawned worker agents is added to the generated Research Report in a markdown format \cite{markdownguide} and once the research is done, the report is saved in a markdown (\textbf{.md}) file in the storage.

\section{Static - Deep Research Agent}
\subsection{Design Overview}

We discuss the technical design of the deep research agent. In this section we talk about the \textbf{Supervisor} agent, the \textbf{Independent} agent and the \textbf{Worker} agent. We also talk about how the agent orchestrates all the above discussed agents to research for a topic and how the agent consolidates the output into the final research report.

At the end we discuss how we can configure the Deep Research Agent with the two parameters “\textbf{Depth}” and “\textbf{Breadth}” to control how in-depth we want our Deep Research Agent to research a given topic. We should be mindful that configuring these parameters to a lower value can reduce the cost (for interaction with LLM) but might also lead to a lower quality research report and vice-versa.

\subsection{Core Design and Parameter Tuning}

The core design of the Deep Research Agent is incorporated in its hierarchical architecture which is controlled by two parameters \textbf{depth} and \textbf{breadth}.

\begin{enumerate}
    \item \textbf{Depth}: The depth parameter controls how deep our agent can research for a provided topic. It also depends upon whether a sub-topic can be further broken into multiple child sub-topics or not.
   \item \textbf{Breadth}: The breadth parameter controls how many sub-topics our agent can break a research topic into. Tuning the breadth parameter into a decent number allows the Large Language Model to research for a topic which is atomic and focussed on a single topic or goal to research for.
\end{enumerate}

Figure 1 demonstrates, when we provide our Deep Research Agent a research topic to research for, the agent aims to divide the topic into multiple sub-topics which can be researched independently. The number of sub-topics spawned from the parent research topic depends upon the breadth parameter. Higher the breadth parameter, greater the number of sub-topics being spawned.

The number of sub-topics being spawned from the parent research topic not only depends on the breadth parameter but also on the limitation of the topic to be broken down into independent researchable sub-topics. If a Research Topic says \textbf{RT} can be broken down only into \(ST_{max}\) independent sub-topics, then the amount of sub-topics spawned from a parent topic can be represented as.

\begin{center}
\(numberOfSubTopics(RT) = min(ST_{max} , b)\)
\end{center}

In Figure 2 we demonstrate for a research topic “What are the investment philosophies of Duan Yongping, Warren Buffett and Charlie Munger” the sub-topic limit is (\(ST_{max}\)) 3. Hence, even though the value of \textbf{breadth} parameter is \textbf{5}, the agent breaks the topic into \textbf{3} independent sub-topics [What are the investment philosophies of Duan Yongping, What are the investment philosophies of  Warren Buffett, What are the investment philosophies of Charlie Munger].

\begin{figure}[H]
    \centering
    \includegraphics[width=0.7\linewidth]{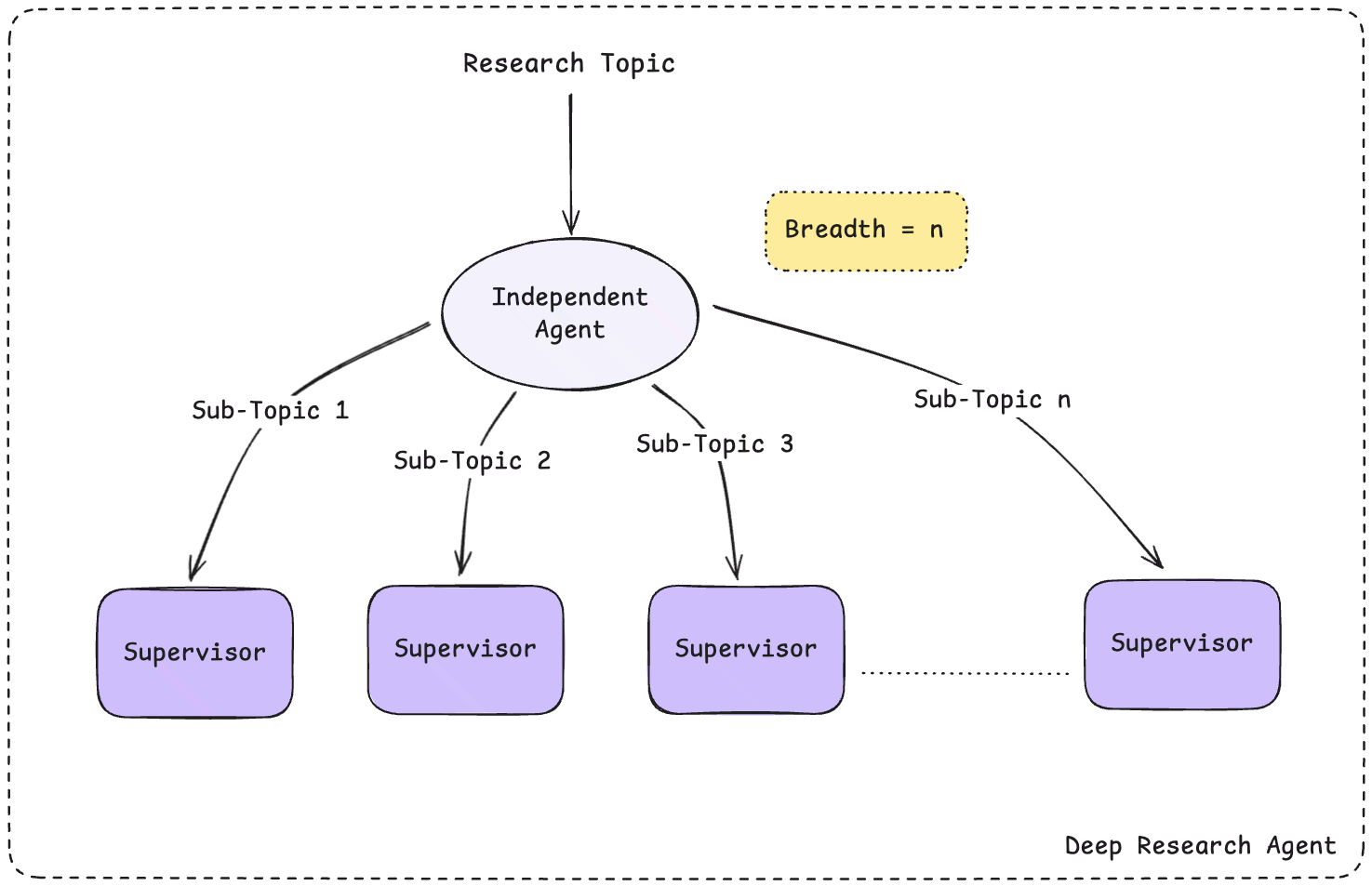}
    \caption{Independent agent breaking down the research topic into sub-topics.}
    \label{fig:placeholder}
\end{figure}

\begin{figure}[H]
    \centering
    \includegraphics[width=0.7\linewidth]{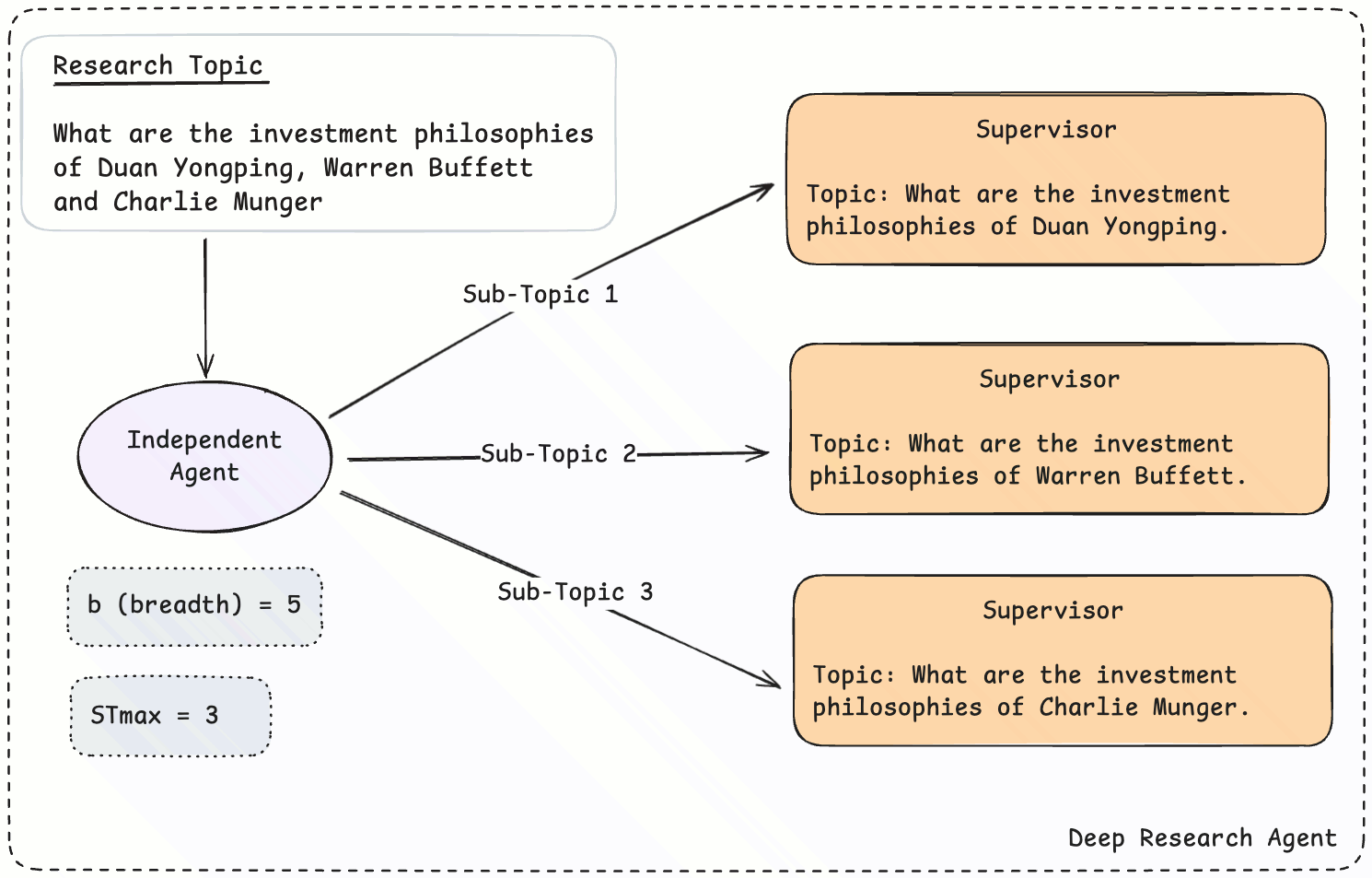}
    \caption{Independent agent breaking down the research topic into maximum possible sub-topics irrespective of the value of breadth (b) parameter}
    \label{fig:placeholder}
\end{figure}

Our agent aims to research by going into the depth of the provided research topic. This behaviour can be configured with the depth parameter. The depth parameter lets the agent break a topic into sub-topics. If the agent reaches the depth limit, then it directly researches that topic, without breaking them further down into sub-topics.

In Figure 3, we describe for a given Research Topic how we can control how deep the agent can go into it. For the research topic “What are the investment philosophies of Duan Yongping, Warren Buffett and Charlie Munger” and the \textbf{depth} parameter value \textbf{3}, the agent went 3 levels deep researching into [“What are the investment philosophies of Warren Buffett”, “What are the core tenets of Warren Buffett's investment philosophy?”, “Define and explain Warren Buffett's concept of the 'circle of competence'”]. Once the agent reached the maximum depth value, it went ahead and researched for that topic without breaking it further into more independent sub-topics.

\begin{figure}[H]
    \centering
    \includegraphics[width=0.7\linewidth]{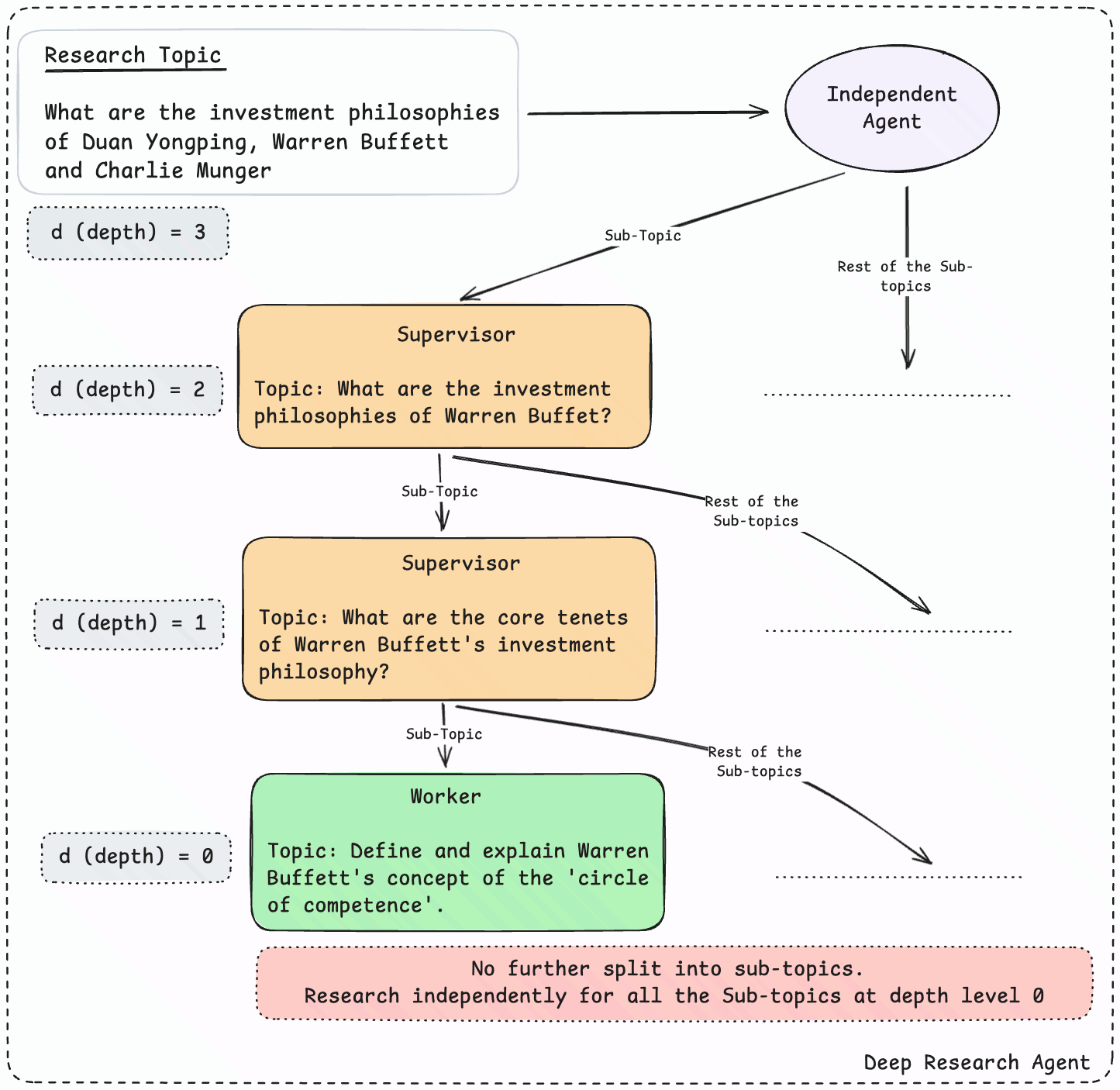}
    \caption{The depth of the research being controlled by the depth (d) parameter.}
    \label{fig:placeholder}
\end{figure}

The overall research procedure of the Deep Research Agent is demonstrated in Table 1 where the parameters \textbf{depth} = \textbf{2}, \textbf{breadth} = \textbf{5} and model \textbf{gemini-2.5-pro} are configured for the agent execution. The table explains how the agent breaks down the topic into independent sub-topics and researches them separately while respecting the configured breadth and the depth limit.

We also aim to reduce the configured \textbf{breadth} value by a factor of \textbf{2}, every time the agent researches a level (depth) deeper. This design decision has been taken to allow the agent to explore the more relevant topics in-detail by breaking them into greater independent sub-queries while the lower level topics can be less relevant and need not to be researched in a huge breadth (detail).

\begin{table}[H] 
    \centering 
    \caption{Static Deep Research Agent (gemini-2.5-pro) [depth = 2, breadth = 5]} 
    \label{tab:my_table} 
    
    \begin{tabular}{| p{5cm} | p{5cm} | p{5cm} |} 
     \hline
     \textbf{Supervisor 1: What are the core tenets of Warren Buffett's investment philosophy, including his concepts of 'circle of competence,' 'economic moats,' and 'margin of safety'?} & 
     \textbf{Supervisor 2: What are the key elements of Charlie Munger's investment philosophy, particularly his emphasis on 'mental models' and the psychology of human misjudgment?} & 
     \textbf{Supervisor 3: What is Duan Yongping's investment philosophy, and how has it been influenced by Buffett and Munger? Focus on his specific principles like 'Benfen' and his approach to technology and consumer electronics investments.} \\ 
     \hline
     \textbf{Worker 1.1}: Define and explain Warren Buffett's concept of the 'circle of competence', detailing its importance in his investment decision-making process and how he advises investors to develop and adhere to their own.
\newline \newline
\textbf{Worker 1.2}: Analyze Warren Buffett's investment principle of 'economic moats', describing the different types of moats he identifies (e.g., brand, patent, switching costs, cost advantages) and providing examples of companies he has invested in that exemplify this concept.
\newline \newline
\textbf{Worker 1.3}: Investigate Warren Buffett's 'margin of safety' principle, explaining how it is defined, its role in minimizing downside risk, and its relationship to the calculation of a company's intrinsic value.  & 
     \textbf{Worker 2.1}: Detail Charlie Munger's core investment philosophy, focusing on his principles of value investing, patience, and the importance of investing in high-quality businesses. Exclude his concepts of 'mental models' and 'psychology of human misjudgment' from this query.
\newline \newline
\textbf{Worker 2.2}: Investigate Charlie Munger's concept of 'mental models'. Define what he meant by a 'latticework of mental models' and provide a comprehensive list and explanation of the key models he advocated for, drawing from disciplines like physics, biology, and economics.
\newline \newline
\textbf{Worker 2.3}: Explore Charlie Munger's 'Psychology of Human Misjudgment'. Identify and explain the primary cognitive biases and psychological tendencies he believed lead to poor decision-making in investing, such as confirmation bias, loss aversion, and social proof. & 
     \textbf{Worker 3.1}: Detail Duan Yongping's core investment philosophy, with a primary focus on his principle of 'Benfen'. Explain what 'Benfen' means and how it translates into specific, actionable investment criteria he follows, such as focusing on business models, management quality, and long-term value over short-term market fluctuations.
\newline \newline
\textbf{Worker 3.2}: Analyze the specific influences of Warren Buffett and Charlie Munger on Duan Yongping's investment approach. Identify key concepts he adopted, such as 'circle of competence', 'margin of safety', and the importance of investing in great businesses at fair prices. Provide examples of how he has adapted and integrated these concepts with his own unique perspective.
\newline \newline
\textbf{Worker 3.3}: Investigate Duan Yongping's investment strategy and major decisions within the technology and consumer electronics sectors. Analyze his significant investments, such as Apple, and explain how these choices exemplify the application of his 'Benfen' principle and Buffett-Munger-inspired philosophy. Contrast his approach with typical venture capital or momentum investing in the tech industry. \\
     \hline
    \end{tabular}

\end{table}

With our Deep Research Agent we can decide on the amount of depth and breadth by which we want our agent to research on a given topic. We ran our Deep Research Agent for a research topic “What are the investment philosophies of Duan Yongping, Warren Buffett, and Charlie Munger?” at three configurations of breadth and depth parameters: \textbf{agent\_d1\_b2} with depth value \textbf{1} and breadth value \textbf{2}, \textbf{agent\_d2\_b3} with depth value \textbf{2} and breadth value \textbf{3}, \textbf{agent\_d2\_b5} with depth value \textbf{2} and breadth value \textbf{5}. The variations in the size of the generated research report, number of research sub-topics generated, and the scores (evaluated via \textbf{DeepResearch Bench} \cite{du2025deepresearchbenchcomprehensivebenchmark}) is presented in Figure 4. The naming convention of the agent configuration follows “\textbf{agent\_dx\_by}” format where \textbf{x} is the depth value and \textbf{y} is the breadth value.

\begin{figure}[H]
    \centering
    \includegraphics[width=1\linewidth]{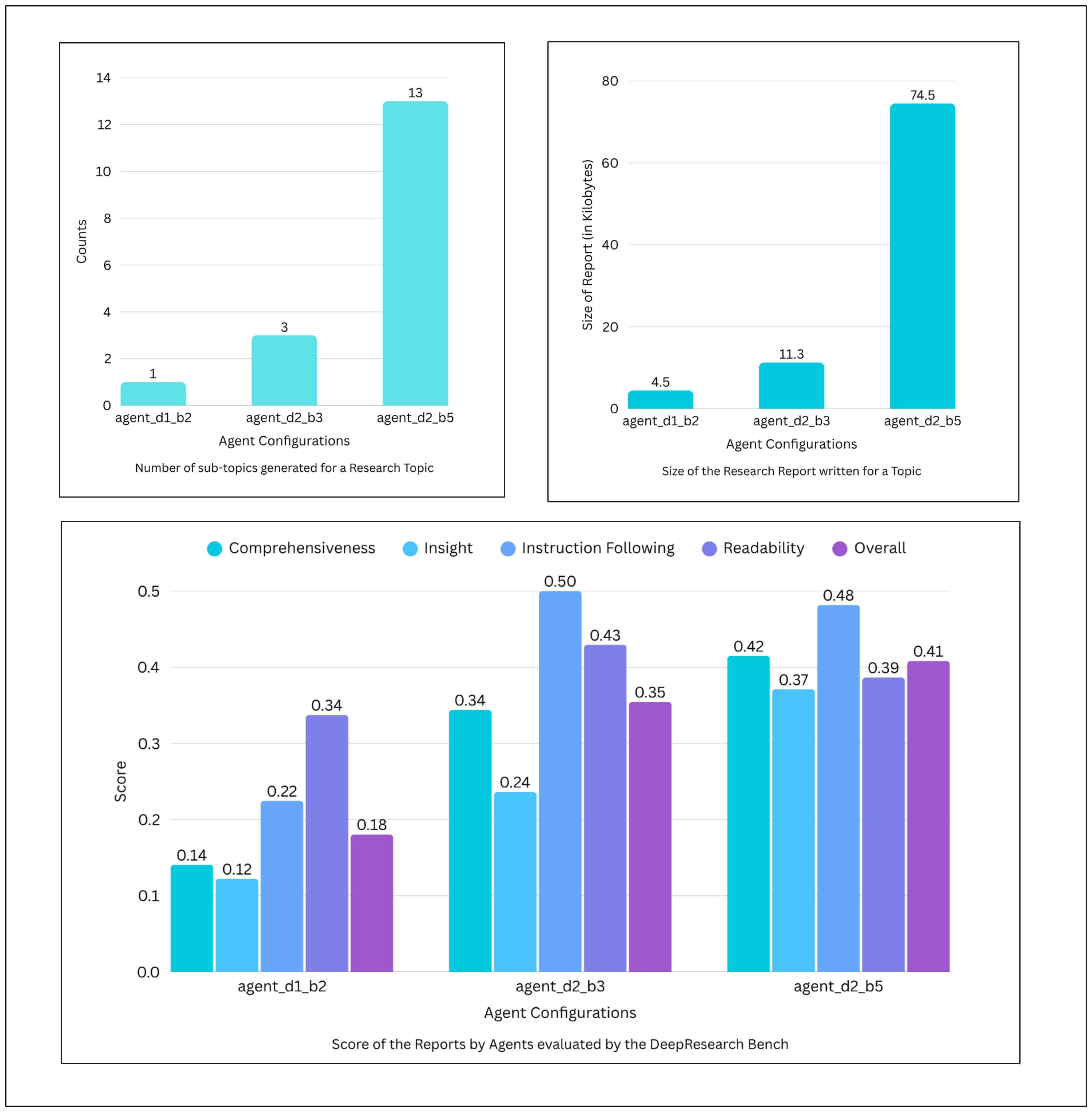}
    \caption{Evaluation Metrics for multiple agent configurations.}
    \label{fig:placeholder}
\end{figure}

We show that with the increasing value of the depth (d) and breadth (b) parameters, the deep research agent does a more in-depth research on the provided topic and hence results in a greater overall score. We also present that with the changing value of depth and breadth configurations, the capability of the Large Language Model to split a research topic into sub-topics changes and the overall depth of the research process also changes. This is demonstrated in Figure 5.

\begin{figure}[H]
    \centering
    \includegraphics[width=1\linewidth]{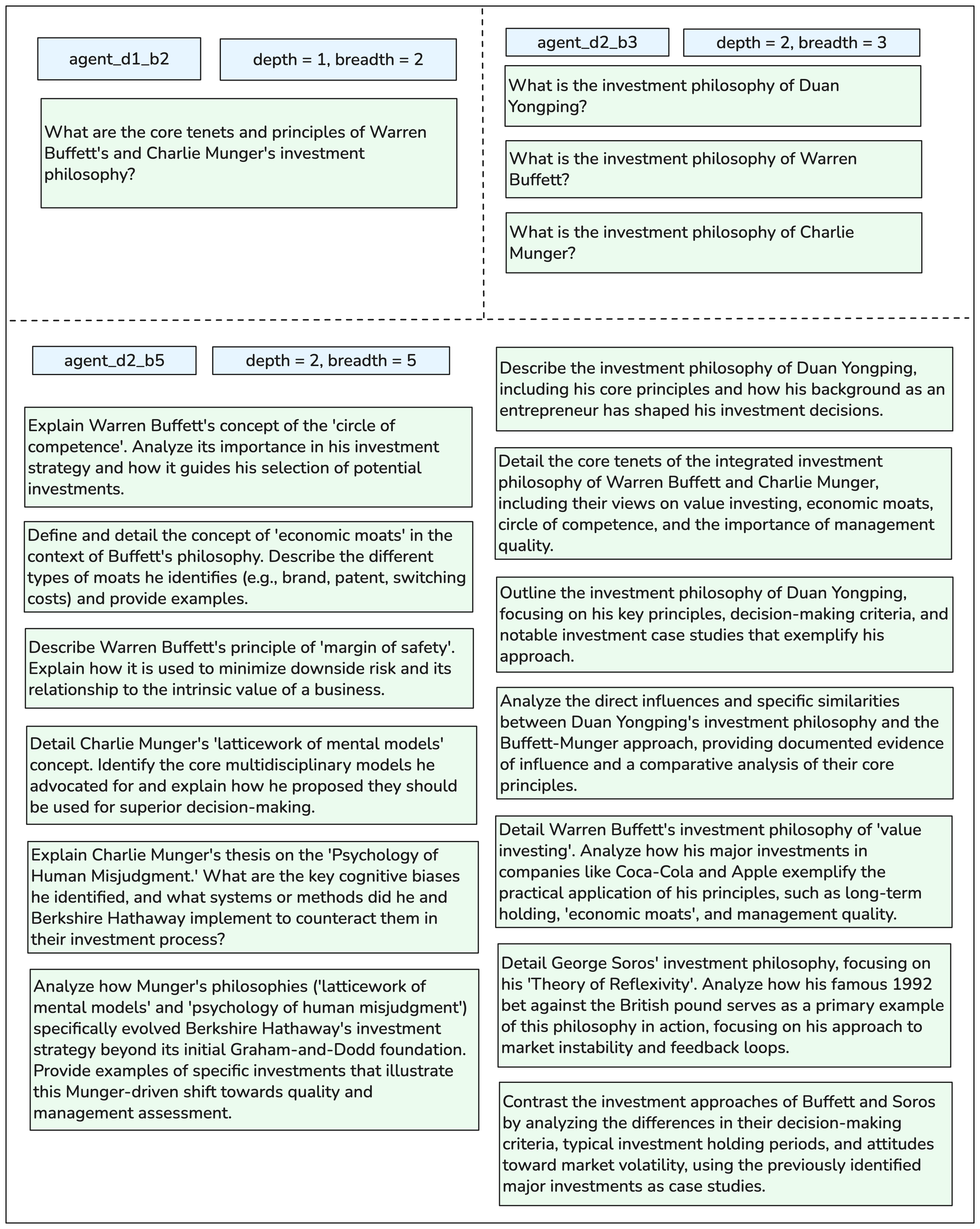}
    \caption{Increase in the ability to perform in-depth research with increasing value of depth (b) and breadth (b) parameters.}
    \label{fig:placeholder}
\end{figure}

\subsection{Supervisor Agent}

The algorithm in Algorithm 1 outlines how a Supervisor Agent makes a decision on whether to split a research topic into independent sub-topics or not. The supervisor agent also takes the configured depth parameter into consideration while deciding on splitting the query. The agent interacts with a Large Language Model to decide whether a query is splittable or not. The algorithm mentioned lays out how the Supervisor agent executes. If the depth parameter has not been exhausted and the research topic can be split into sub-topics, the agent forwards the topic to the Independent agent.

\begin{algorithm}[h]
    \SetAlgoLined
    \caption{Supervisor Agent}
    \label{algo:merge}
    
    \KwData{str $researchTopic$, integer $currentDepth$}
    \KwResult{Shared State $state$}

    $canSplit \leftarrow false$\;

    \If{$currentDepth > 0$}{
        $canSplit \leftarrow canSplitIntoSubtasks(researchTopic)$\;
    }

    \eIf{$canSplit$}{
        \Return $independentAgent(researchTopic)$\;
    }{
        $isUniqueTopic \leftarrow isDifferentResearchTopic(researchTopic)$\;
        \If{$isUniqueTopic$}{
            \Return $worker(researchTopic)$\;
        }
    }
\end{algorithm}

If the research topic can not be split into sub-tasks or the depth parameter has been exhausted, the agent gives the topic to a worker which is responsible for researching the topic with the help of LLM and Web Search tool. Before executing a worker agent, we do a sanity check if the research topic being given to be researched on is not equivalent to the previously searched topics. If it is equivalent, we skip the worker execution for that topic. This sanity check is carried out by the \textbf{isDifferentResearchTopic} method and requires an LLM interaction.

\subsection{Independent Agent}

The algorithm in Algorithm 2 demonstrates how an Independent Agent takes a research topic and breaks them down into sub-topics which can be researched independently. It spawns 1 Supervisor Agent for every sub-topic. The agent acts as a parent in the hierarchy which spawns multiple child supervisors which research for the sub-tasks independently and return their findings back to the parent agent.

\begin{algorithm}[h]
    \SetAlgoLined
    \caption{Independent Agent}
    \label{algo:merge}
    
    \KwData{integer $currentDepth$, integer $currentBreadth$}
    \KwResult{Shared State $state$}
    
    \ForEach{$subQuery \in independentAgentResponse.subQueries$}{    
        
        $supervisorResponse \leftarrow supervisor(
            subQuery, 
            currentDepth - 1, 
            currentBreadth - 2
        )$\;
        
        $supervisorResponses.append(supervisorResponse)$\;
    }
\end{algorithm}

The algorithm in Algorithm 3 demonstrates how a parent independent agent then collects the output of all the child supervisors and combines them into a unified Research Report response which is returned back to the caller.

\begin{algorithm}[h]
    \SetAlgoLined
    \caption{Independent Agent}
    \label{algo:merge}
    
    \KwData{}
    \KwResult{Shared State $state$}
    \ForEach{$supervisorResponse \in supervisorResponses$}{
        $pastResearchReports.extend(supervisorResponse[\texttt{"reports"}])$\;
        
        $pastCitations.extend(supervisorResponse[\texttt{"citations"}]$)\;
        
        $pastResearchTopics.extend(supervisorResponse[\texttt{"research\_topics"}]$)\;
    }
\end{algorithm}

The Research Report holds these important informations:

\begin{enumerate}
    \item \textbf{Current Depth}: The current depth value for the research agent. The current depth value is reduced by a factor of 1 every time an Independent agent passes the sub-query to the Supervisor agent.
    \item \textbf{Current Breadth}: The current breadth value for the research agent. The current breadth value is reduced by a factor of 2 every time an Independent agent passes the sub-query to the Supervisor agent. 
    \item \textbf{Current Research Topic}: The current research topic to be researched on.
    \item \textbf{Past Research Topics}: The past research topics which are already being researched from the start of the deep research agent execution.
    \item \textbf{Past Citations}: The past citations collected through the web search while researching on the past research topics from the start of the deep research agent execution.
    \item \textbf{Past Research Reports}: The past research reports which are already being generated from the start of the deep research agent execution.
\end{enumerate}

The process of unifying the research responses from the child supervisors by the parent independent agent involves extending the citations, reports and the topics list. The independent agent has a list of citations, reports and topics which has been collected by the deep research agent execution till that instant. It further adds the elements in the citations, reports and the topics list returned by every child supervisor into the currently maintained list.

\subsection{Worker Node \& Web Search Tool}
\subsubsection{Worker Node}
A worker node is responsible for researching the topic provided to it with the help of a Web Search tool and Large Language Model. The task of a worker node is to perform an in-depth research on a provided research topic (by the supervisor). The worker node performs their research in two parts.

\subsubsection{ Calling a Web Search Tool}
The worker node has been provided with a tool calling capability which allows them to search the web for a given research topic. We use \textbf{Tavily} \cite{tavilywebsearch} for web search and aim to receive \textbf{top 5} search results for a topic from the web.  We also make sure to have only relevant web search results with us for writing a report for the research topic. The Tavily Web Search tool provides a score field for every search result returned which defines “the relevance score of the search result”. We have a \textbf{threshold score} value set to \textbf{30\%} which filters out any search result whose score is less than the threshold score.

We also keep the \textbf{cited urls} of the relevant search responses for report writing purposes. The search response has a url field which holds the URL of the search result. The structure of a single search response being returned from the Web Search tool is mentioned in Listing 1. A deep dive into the documentation of the web search API can be found on their documentation \cite{tavilydocs} website.
\\

\begin{lstlisting}[language=json, caption={Web Search Response from Tavily}]
{
    "title": "Lionel Messi Facts | Britannica",
    "url": "https://www.britannica.com/facts/Lionel-Messi",
    "content": "Lionel Messi, an Argentine footballer...",
    "score": 0.81025416,
    "raw_content": null,
    "favicon": "https://britannica.com/favicon.png"
}
\end{lstlisting}

\subsubsection{Researching the Topic with LLM}
Once we have the Web Search response (a collection of most relevant search results), we provide that search response to a Large Language Model along with the research topic (currently being researched) and a structured prompt (holding detailed instructions on the research process) and let them perform an in-depth research of the topic. We have used “\textbf{gemini-2.5-pro}” LLM for this deep research agent and the entire evaluation scores are based on the same model.

The deep research agent allows an easy switching between the large language models for report writing. If in future a more capable LLM is released, the current deep research agent can be updated to utilize that model capabilities very efficiently and with ease.

The model’s generated output is added to the generated past research reports list and is further utilized for the full report writing. Along with the model output we also maintain a list of citations which is a collection of the urls retrieved from as a part of the most relevant search results. These citations are preserved in the past citations list throughout the deep research agent runtime and are utilized for final report writing.

\subsection{Generated Report structure}
The generated report by the Deep Research Agent is divided into three parts: Table of Contents, Report and Citations. The “Table of Contents” holds the topics researched by the worker agent in the order. Meaning a topic being researched earlier will come first in the table of contents section. The “Report” section holds the main content of the Research Report. The research done by the worker agent is written in this section. This section maintains the order and the hierarchy of the research topics and sub-topics. The “Citation” section holds the list of cited URLs we received from the Web Search tool (in the search response). Section \ref{sec:samplereport} outlines a sample research report generated on a research topic mentioned with the configuration of \(depth = 2\) and \(breadth = 3\).

\section{Evaluation}
We evaluate our Deep Research Agent against the globally accepted benchmark named DeepResearch Bench \cite{du2025deepresearchbenchcomprehensivebenchmark}\cite{deepresearchbenchpage}. It is the primary benchmark for general-purpose DRAs. It consists of 100 PhD-level research tasks across 22 distinct fields. Crucially, it also introduces two novel evaluation frameworks for grading the results:

\begin{enumerate}
    \item \textbf{RACE} (Reference-based Adaptive Criteria-driven Evaluation): Assesses the quality of the final generated research report.
    \item \textbf{FACT} (Framework for Factual Abundance and Citation Trustworthiness): Assesses the agent's retrieval and citation capabilities.
\end{enumerate}

We evaluated our deep research agent against the RACE evaluation framework of the DeepResearch Bench and with the configuration of \textbf{depth} (d) value \textbf{2} and \textbf{breadth} (b) value \textbf{5} and \textbf{gemini-2.5-pro} model powering the supervisor and worker agents, our deep research agent was able to achieve an overall score of \textbf{34.72}. The detailed evaluation is mentioned in Table 2. The evaluation scores of the models have been captured from the DeepResearch Bench Hugging Face leaderboard \cite{deepresearchleaderboard}.

\begin{table}[H] 
    \centering 
    \caption{RACE (Reference-based Adaptive Criteria-driven Evaluation) by models} 
    \label{tab:my_table} 
    
    \begin{tabular}{| p{4.5cm} | p{1.5cm} | p{3cm} |  p{1.2cm} |  p{2cm} |  p{2cm} |} 
     \hline
     \textbf{Model} & 
     \textbf{Overall} & 
     \textbf{Comprehensive- ness} &
     \textbf{Insight} &
     \textbf{Instruction Following} &
     \textbf{Readability} \\ 
     \hline
     gemini-2.5-pro-deepresearch & 49.71 & 49.51 & 49.45 & 50.12 & 50  \\
     \hline
     openai-deep-research & 46.45 & 46.46 & 43.73 & 49.39 & 47.22 \\
     \hline
     claude-research & 45 & 45.34 & 42.79 & 47.58 & 44.66 \\
     \hline 
     static-dra (gemini-2.5-pro) & 34.72 & 35.12 & 30.45 & 38.86 & 35.44 \\
     \hline
     gemini-2.5-pro-preview-05-06 & 31.9 & 31.75 & 24.61 & 40.24 & 32.76 \\
     \hline
     gpt-4o-search-preview & 30.74 & 27.81 & 20.44 & 41.01 & 37.6 \\
     \hline
     sonar & 30.64 & 27.14 & 21.62 & 40.7 & 37.46 \\
     \hline
    \end{tabular}
\end{table}

The scores of our deep research agent evaluated on the DeepResearch Bench for \textbf{English} and \textbf{Chinese} are denoted in Table 3.

\begin{table}[H] 
    \centering 
    \caption{RACE (Reference-based Adaptive Criteria-driven Evaluation) by languages} 
    \label{tab:my_table} 
    
    \begin{tabular}{| p{3.5cm} | p{1.5cm} | p{3cm} |  p{1.5cm} |  p{2cm} |  p{2cm} |} 
     \hline
     \textbf{Model (Language)} & 
     \textbf{Overall} & 
     \textbf{Comprehensive- ness} &
     \textbf{Insight} &
     \textbf{Instruction Following} &
     \textbf{Readability} \\ 
     \hline
     static-dra (english) & 32.65 & 33.23 & 26.65 & 38.20 & 33.87 \\
     \hline
     static-dra (chinese) & 36.80 & 37.01 & 34.25 & 39.52 & 37.02 \\
     \hline
    \end{tabular}
\end{table}

The DeepResearch Bench has multiple topics across which we evaluated our deep research agent and the evaluation metrics across these topics are provided in Table 4. A heatmap of the evaluation scores against all the supported topics is depicted in Figure 6.

\begin{table}[H] 
    \centering 
    \caption{RACE (Reference-based Adaptive Criteria-driven Evaluation) by topics} 
    \label{tab:my_table} 
    
    \begin{tabular}{| p{3.5cm} | p{1.5cm} | p{3cm} |  p{1.5cm} |  p{2cm} |  p{2cm} |} 
     \hline
     \textbf{static-dra (Topic)} & 
     \textbf{Overall} & 
     \textbf{Comprehensive- ness} &
     \textbf{Insight} &
     \textbf{Instruction Following} &
     \textbf{Readability} \\ 
     \hline
     Finance \& Business & 32.65 & 33.96 & 28.35 & 36.73 & 33.04 \\
     \hline
     Science \& Technology & 32.22 & 32.68 & 29.28 & 35.07 & 33.54 \\
     \hline
     Software Development & 33.71 & 34.82 & 28.78 & 37.16 & 35.41 \\
     \hline
     Education \& Jobs & 35.68 & 35.93 & 30.90 & 40.59 & 36.50 \\
     \hline
     Health & 39.22 & 38.83 & 34.69 & 45.93 & 37.90 \\
     \hline
     Literature & 31.16 & 29.54 & 23.43 & 39.09 & 33.24 \\
     \hline
     History & 39.60 & 40.89 & 36.55 & 41.70 & 40.28 \\
     \hline
     Hardware & 37.81 & 37.88 & 30.43 & 43.03 & 38.99 \\
     \hline
     Industrial & 34.58 & 36.55 & 29.73 & 37.52 & 36.15 \\
     \hline
     Art \& Design & 34.70 & 36.72 & 30.03 & 38.71 & 34.92 \\
     \hline
     Games & 29.17 & 25.05 & 26.08 & 27.13 & 44.53 \\
     \hline
     Crime \& Law & 34.46 & 34.42 & 30.62 & 40.94 & 31.86 \\
     \hline
     Entertainment & 29.77 & 25.33 & 23.04 & 32.11 & 33.26 \\
     \hline
     Sports \& Fitness & 34.72 & 31.25 & 32.73 & 42.14 & 30.70 \\
     \hline
     Software & 30.44 & 30.52 & 25.46 & 35.38 & 33.47 \\
     \hline
     Transportation & 35.93 & 34.70 & 33.18 & 39.88 & 35.75 \\
     \hline
     Religion & 37.08 & 36.45 & 36.37 & 41.08 & 35.28 \\
     \hline
     Home \& Hobbies & 35.82 & 38.20 & 32.91 & 41.56 & 29.24 \\
     \hline
     Travel & 37.10 & 41.17 & 31.68 & 39.93 & 35.51 \\
     \hline
     Food \& Dining & 39.40 & 39.42 & 34.23 & 41.69 & 45.13 \\
     \hline
     Fashion \& Beauty & 37.88 & 37.61 & 35.05 & 43.33 & 35.25 \\
     \hline
     Social Life & 44.21 & 46.21 & 41.75 & 48.23 & 40.72 \\
     \hline
    \end{tabular}
\end{table}

\begin{figure}[H]
    \centering
    \includegraphics[width=1.0\linewidth]{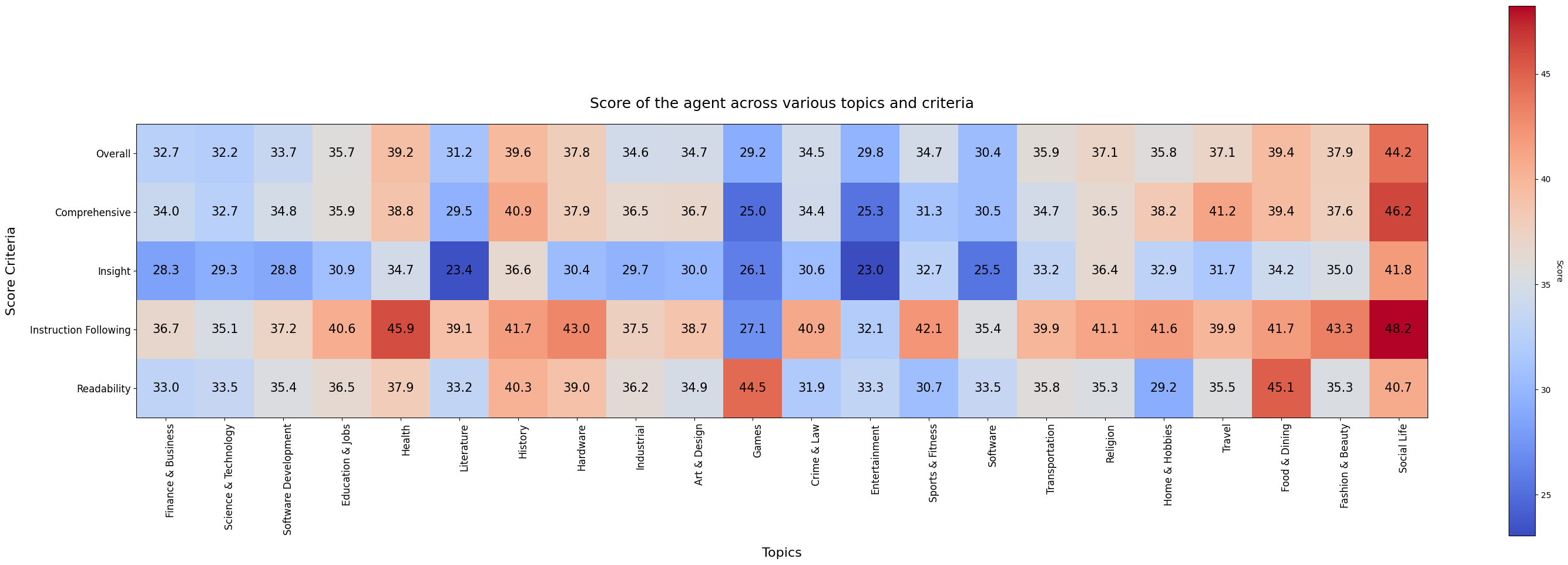}
    \caption{Score of the Agent across various topics and criteria}
    \label{fig:placeholder}
\end{figure}

\section{Conclusion}

The Static Deep Research Agent (Static-DRA) presented in this paper introduces a novel, configurable, and hierarchical tree-based architecture for tackling complex, multi-turn research tasks. Our core contribution is the design of a static workflow governed by two user-tunable parameters: Depth and Breadth. These parameters provide a critical mechanism for end-users to exert granular control over the extent of the research process, directly balancing the desired quality and comprehensiveness of the report against the computational cost associated with Large Language Model (LLM) interactions.

The agent's hierarchical design, composed of Supervisor, Independent, and Worker agents, enables effective multi-hop information retrieval and parallel sub-topic investigation. The dynamic reduction of the Breadth parameter at deeper levels ensures that the agent intelligently focuses its resources on the most promising research avenues.

Our evaluation against the established DeepResearch Bench using the RACE framework validates the agent's capability. The Static-DRA, configured with a depth of 2 and breadth of 5, achieved an overall score of 34.72, demonstrating competitive performance powered by the gemini-2.5-pro model. Furthermore, our experiments confirm the fundamental trade-off: increasing the configured Depth and Breadth values results in a more in-depth research process and a correspondingly higher evaluation score and report size. In conclusion, the Static-DRA provides a pragmatic and resource-aware solution for deep research, giving users transparent control over the research intensity.

\bibliographystyle{plain} 
\bibliography{references}

@misc{huang2025deepresearchagentssystematic,
      title={Deep Research Agents: A Systematic Examination And Roadmap}, 
      author={Yuxuan Huang and Yihang Chen and Haozheng Zhang and Kang Li and Huichi Zhou and Meng Fang and Linyi Yang and Xiaoguang Li and Lifeng Shang and Songcen Xu and Jianye Hao and Kun Shao and Jun Wang},
      year={2025},
      eprint={2506.18096},
      archivePrefix={arXiv},
      primaryClass={cs.AI},
      url={https://arxiv.org/abs/2506.18096}, 
}

@misc{yan2024correctiveretrievalaugmentedgeneration,
      title={Corrective Retrieval Augmented Generation}, 
      author={Shi-Qi Yan and Jia-Chen Gu and Yun Zhu and Zhen-Hua Ling},
      year={2024},
      eprint={2401.15884},
      archivePrefix={arXiv},
      primaryClass={cs.CL},
      url={https://arxiv.org/abs/2401.15884}, 
}

@misc{selfretrievalaugmentedgeneration,
      title={Self-RAG: Learning to Retrieve, Generate, and Critique through Self-Reflection}, 
      author={Akari Asai and Zeqiu Wu and Yizhong Wang and Avirup Sil and Hannaneh Hajishirzi},
      year={2023},
      eprint={2310.11511},
      archivePrefix={arXiv},
      primaryClass={cs.CL},
      url={https://arxiv.org/abs/2310.11511}, 
}

@misc{zhang2025deepresearchsurveyautonomous,
      title={Deep Research: A Survey of Autonomous Research Agents}, 
      author={Wenlin Zhang and Xiaopeng Li and Yingyi Zhang and Pengyue Jia and Yichao Wang and Huifeng Guo and Yong Liu and Xiangyu Zhao},
      year={2025},
      eprint={2508.12752},
      archivePrefix={arXiv},
      primaryClass={cs.IR},
      url={https://arxiv.org/abs/2508.12752}, 
}

@misc{lewis2021retrievalaugmentedgenerationknowledgeintensivenlp,
      title={Retrieval-Augmented Generation for Knowledge-Intensive NLP Tasks}, 
      author={Patrick Lewis and Ethan Perez and Aleksandra Piktus and Fabio Petroni and Vladimir Karpukhin and Naman Goyal and Heinrich Küttler and Mike Lewis and Wen-tau Yih and Tim Rocktäschel and Sebastian Riedel and Douwe Kiela},
      year={2021},
      eprint={2005.11401},
      archivePrefix={arXiv},
      primaryClass={cs.CL},
      url={https://arxiv.org/abs/2005.11401}, 
}

@misc{openaideepresearch,
  author       = {OpenAI},
  title        = {Introducing deep research},
  howpublished = {\url{https://openai.com/index/introducing-deep-research/}},
  year         = {2025},
  note         = {Accessed: 2025-02-03}
}

@misc{geminideepresearch,
  author       = {Google},
  title        = {Gemini Deep Research},
  howpublished = {\url{https://gemini.google/overview/deep-research/}},
}

@misc{xaideepresearch,
  author       = {xAI Team},
  title        = {Grok 3 Beta — The Age of Reasoning Agents},
  howpublished = {\url{https://x.ai/news/grok-3}},
  year         = {2025},
  note         = {Accessed: 2025-02-19}
}

@misc{perplexitydeepresearch,
  author       = {Perplexity Team},
  title        = {Introducing Perplexity Deep Research},
  howpublished = {\url{https://www.perplexity.ai/hub/blog/introducing-perplexity-deep-research}},
  year         = {2025},
  note         = {Accessed: 2025-02-14}
}

@misc{langchaindeepresearch,
  author       = {LangChain Team},
  title        = {Open Deep Research},
  howpublished = {\url{https://github.com/langchain-ai/open_deep_research}},
  year         = {2025},
  note         = {Accessed: 2025-07-16}
}

@misc{zhang2024enhancinglargelanguagemodel,
      title={Enhancing Large Language Model Performance To Answer Questions and Extract Information More Accurately}, 
      author={Liang Zhang and Katherine Jijo and Spurthi Setty and Eden Chung and Fatima Javid and Natan Vidra and Tommy Clifford},
      year={2024},
      eprint={2402.01722},
      archivePrefix={arXiv},
      primaryClass={cs.CL},
      url={https://arxiv.org/abs/2402.01722}, 
}

@misc{zhang2023iaginductionaugmentedgenerationframework,
      title={IAG: Induction-Augmented Generation Framework for Answering Reasoning Questions}, 
      author={Zhebin Zhang and Xinyu Zhang and Yuanhang Ren and Saijiang Shi and Meng Han and Yongkang Wu and Ruofei Lai and Zhao Cao},
      year={2023},
      eprint={2311.18397},
      archivePrefix={arXiv},
      primaryClass={cs.CL},
      url={https://arxiv.org/abs/2311.18397}, 
}

@misc{luo2023sailsearchaugmentedinstructionlearning,
      title={SAIL: Search-Augmented Instruction Learning}, 
      author={Hongyin Luo and Yung-Sung Chuang and Yuan Gong and Tianhua Zhang and Yoon Kim and Xixin Wu and Danny Fox and Helen Meng and James Glass},
      year={2023},
      eprint={2305.15225},
      archivePrefix={arXiv},
      primaryClass={cs.CL},
      url={https://arxiv.org/abs/2305.15225}, 
}

@misc{markdownguide,
  author       = {The Markdown Team},
  title        = {Markdown Guide},
  howpublished = {\url{https://www.markdownguide.org/}},
}

@misc{tavilywebsearch,
  author       = {Tavily Team},
  title        = {Web Search - Connect Your Agent to the Web},
  howpublished = {\url{https://www.tavily.com/}},
}

@misc{tavilydocs,
  author       = {Tavily Team},
  title        = {Web Search documentation},
  howpublished = {\url{https://docs.tavily.com/documentation/api-reference/endpoint/search}},
}

@misc{du2025deepresearchbenchcomprehensivebenchmark,
      title={DeepResearch Bench: A Comprehensive Benchmark for Deep Research Agents}, 
      author={Mingxuan Du and Benfeng Xu and Chiwei Zhu and Xiaorui Wang and Zhendong Mao},
      year={2025},
      eprint={2506.11763},
      archivePrefix={arXiv},
      primaryClass={cs.CL},
      url={https://arxiv.org/abs/2506.11763}, 
}

@misc{deepresearchbenchpage,
  author       = {DeepResearch Bench Team},
  title        = {DeepResearch Bench: A Comprehensive Benchmark for Deep Research Agents},
  howpublished = {\url{https://deepresearch-bench.github.io/}},
  year={2025},
}

@misc{deepresearchleaderboard,
  author       = {HuggingFace Team},
  title        = {DeepResearch Bench: Leaderboard},
  howpublished = {\url{https://huggingface.co/spaces/muset-ai/DeepResearch-Bench-Leaderboard}},
  year={2025},
}

\section{Computing the number of research topics being provided to the Worker}
By configuring the depth and breadth parameters we can also control the amount of queries being spawned by our Deep Research Agent for a given Research Topic. This can also be used to have a control / limitation on the interactions with the large language model. This can be calculated through the following mathematical steps.
\\ \\
Configured Depth Value = \(d\)
\\
Configured Breadth Value = \(b\)
\\ \\
Let’s assume the number of sub-topics spawned at depth k is ns(k)
\\ \\
Maximum number of subtopics spawned at depth 1 = \(ns(1) = b = (b - 0*2)\)
\\
Maximum number of subtopics spawned at depth 2 = \(ns(1) * (b - 1*2)\)
\\
Maximum number of subtopics spawned at depth 3 = \(ns(2) * (b - 2*2)\)
\\
\(…\)
\\
Maximum number of sub-topics spawned at depth k = \(ns(k-1) * [b - (k-1)*2]\)
\\
Maximum number of sub-topics spawned at depth d = \(ns(d-1) * [b - (d-1)*2]\)
\\ \\
Since the maximum number of sub-topics spawned at the lowest level is provided to the workers for research via interaction with the Large Language models, we can safely say that the maximum number of topics provided to the LLM for research is equivalent to ns(d).
\\ \\
\(ns(d) = ns(d-1) * [b - (d-1)*2]\)
\\
\(ns(d) = ns(d-2) * [b - (d-2)*2] * [b - (d-1)*2]\)
\\
\(ns(d) = b * (b-1*2) * … * [b - (d-2)*2] * [b - (d-1)*2]\)
\\ \\
Hence,
\\
\begin{center}
    \(ns(d) = \prod_{i=0}^{d-1}(max((b-2*i), 1))\) 
\end{center}

\section{Sample Deep Research Report}\label{sec:samplereport}

\subsection{Deep Research Topic}
What are the investment philosophies of Duan Yongping, Warren Buffett, and Charlie Munger?

\subsection{Generated Deep Research Report}

\textbf{Deep Research Report}
\\ \\
\textbf{Table of Contents }
\begin{enumerate}
    \item What is the investment philosophy of Duan Yongping?
    \item What is the investment philosophy of Warren Buffett?
    \item What is the investment philosophy of Charlie Munger?
\end{enumerate}
\textbf{Report} 
\\ \\
\textbf{What are the investment philosophies of Duan Yongping, Warren Buffett, and Charlie Munger?}
\\ \\
What is the investment philosophy of Duan Yongping?
\\ \\
Duan Yongping's investment philosophy is rooted in **value investing**, with a focus on long-term, high-conviction bets in companies he thoroughly understands. His approach is often compared to that of Warren Buffett, emphasizing simplicity and a deep understanding of business fundamentals over chasing trends.
\\ \\
Core Principles of Duan Yongping's Investment Philosophy:
\begin{itemize}
    \item \textbf{Value-Driven and Long-Term Horizon}: Yongping is known for his profound value investment philosophy, seeking out high-quality assets that are undervalued by the market. His decisions are guided by long-term impacts rather than immediate benefits, embodying a "do the right thing" mentality with a long-term vision (https://www.binance.com/en/square/post/18537661287098). This strategy is characterized as "high-conviction, value-driven, long-term investing".
    \item \textbf{Concentrated Portfolio}: Unlike investors who diversify across hundreds of stocks, Yongping prefers a concentrated portfolio, typically holding only 8-12 stocks. This reflects his high-conviction approach, where he invests significantly in a few businesses he believes in (https://www.gainify.io/blog/duan-yongping-portfolio).
    \item \textbf{Deep Understanding Over Chasing Trends}: A cornerstone of his philosophy is to gain a deep, essential understanding of a company's business model, products, and user needs. He advocates for selective learning and avoids blindly chasing new things or imitating trends, a crucial discipline in an era of information overload (https://www.binance.com/en/square/post/18537661287098). His business philosophy also includes being cautious about the risks and uncertainties associated with new technologies and products (https://link.springer.com/content/pdf/10.1007/978-981-95-0545-6\_10.pdf).
    \item \textbf{Adaptability within a Value Framework}: While a disciplined value investor, Yongping demonstrates adaptability. He is willing to embrace growth opportunities, as shown by his investments in companies like Alphabet and NVIDIA, when the long-term economics are favorable. This indicates that his value framework is flexible enough to incorporate growth assets (https://www.gainify.io/blog/duan-yongping-portfolio). 
    \item \textbf{Simplicity and Consistency}: His investment style is consistent with his life philosophy, which centers on simplifying complexities and focusing on the essence (https://www.alphaexponent.net/p/22-duan-the-dilettante, https://www.binance.com/en/square/post/18537661287098). This approach involves focusing on doing one thing well and maintaining a clear, uncluttered mindset.
\end{itemize}
What is the investment philosophy of Warren Buffett?
\\ \\
Warren Buffett's investment philosophy is a disciplined, principle-based approach to investing that has generated significant wealth over several decades (Simply Ethical). It is rooted in the value investing school of Benjamin Graham, but Buffett has expanded upon these initial principles (Investopedia). His strategy serves as a guide for long-term investors, emphasizing wealth creation while avoiding the pitfalls of short-term speculation (Simply Ethical).
\\ \\
The core tenets of his philosophy include:

\begin{itemize}
    \item \textbf{Focus on Value Investing}: At its heart, Buffett's strategy is about value investing. This involves identifying a company's intrinsic value by analyzing its business fundamentals, such as earnings, revenue, and assets (Investopedia). The goal is to purchase these companies when they are undervalued, particularly during market downturns when prices become more attractive (IIFL Capital, Investopedia).
    \item \textbf{Invest in Quality Businesses}: Buffett's philosophy has evolved from just buying cheap stocks to investing in "wonderful businesses at 'fair' valuations" (Investor.fm, HBR). He describes a good business as a "castle" with a protective "moat" that management should continuously widen. These high-quality businesses are often unique franchises that consistently generate cash (HBR). However, finding such businesses that remain good investments over time can be difficult (Investor.fm).
    \item \textbf{Maintain a Long-Term Perspective}: Buffett's principles are designed for long-term success. He invests in a company based on its underlying business quality, not on whether the market will soon recognize its worth (Investopedia). This approach acts as a shield against "the most common pitfalls of speculation and short-termism" (Simply Ethical).
    \item \textbf{Demand Capable and Transparent Management}: The quality and transparency of a company's management are non-negotiable elements in Buffett's approach (Investor.fm). His proposition to managers of the businesses he invests in is that if their company generates cash, they can trust him to reinvest it wisely (HBR).
    \item \textbf{Emphasize Simplicity}: Buffett advocates for a simple investment strategy (IIFL Capital). For investors who may not have the time or expertise to analyze individual businesses, he has recommended periodically investing in an index fund, stating that a "know-nothing investor can actually out-perform most investment professionals" this way (IIFL Capital).
    \item \textbf{Practice Patience and Emotional Discipline}: A key takeaway from Buffett's philosophy is that extraordinary results can be achieved through patience and discipline (Simply Ethical). This involves managing emotions and having the patience to wait for the right opportunities (IIFL Capital). Part of this discipline includes keeping cash reserves available to deploy when market conditions are favorable (IIFL Capital).
\end{itemize}
While timeless, implementing Buffett's philosophy today has its challenges. Finding wonderful businesses at fair prices in the current market is difficult, and Buffett's own massive portfolio puts him at a competitive disadvantage compared to those managing smaller sums (Investor.fm). Nonetheless, his core principles offer clear guidance: focus on quality, maintain a long-term view, and keep the strategy simple (IIFL Capital).
\\ \\
What is the investment philosophy of Charlie Munger?
\\
The Investment Philosophy of Charlie Munger
\\ \\
Charlie Munger, the long-time business partner of Warren Buffett, was a legendary investor whose philosophy was built on a foundation of discipline, rationality, and a deep understanding of business fundamentals. His approach can be summarized as buying great companies at fair prices and holding them for the long term.
\\ \\
Core Tenets of Munger's Philosophy:

\begin{itemize}
    \item \textbf{Focus on Quality Businesses}: The cornerstone of Munger's philosophy is that the "real money is in great companies". His strategy prioritized quality over quantity, seeking out high-quality businesses to invest in.
    \item \textbf{Invest within Your "Circle of Competence"}: Munger was a staunch advocate for only investing in businesses that you can thoroughly understand. He called this principle the "circle of competence" and made it clear that if you don't deeply understand a business, you should not buy its stock (https://www.ruleoneinvesting.com/blog/personal-development/5-charlie-munger-investing-lessons-every-value-investor-should-live-by). This approach isn't about picking stocks, but about buying into businesses you comprehend.
    \item \textbf{Patience and Long-Term Holding}: Munger firmly believed in the long-term holding of quality investments (https://www.investinassets.net/p/10-lessons-from-charlie-munger). His strategy was not focused on short-term market movements but on the long-term growth and value creation of the companies in his portfolio.
    \item \textbf{Concentrated Portfolio}: Unlike the common advice to diversify broadly, Munger preferred a non-diversified, concentrated portfolio (https://www.fool.com/investing/how-to-invest/famous-investors/charlie-munger/). This reflects his confidence in making a few high-conviction bets on excellent companies, often suggesting a portfolio of 10-20 high-quality compounders (https://www.investinassets.net/p/10-lessons-from-charlie-munger).
    \item \textbf{Rationality and Discipline}: A key element of his success was an "unmatched rationality and discipline" that he maintained through numerous market cycles (https://www.investinassets.net/p/10-lessons-from-charlie-munger). His philosophy emphasized staying calm during market volatility and steering clear of common behavioral mistakes that trap many investors.
    \item \textbf{Capitalizing on Market Inefficiency}: Munger's approach was rooted in capitalizing on market inefficiencies (https://finance.yahoo.com/news/charlie-mungers-timeless-investment-advice-113042652.html). This involves identifying opportunities where the market has mispriced a great business, allowing for the purchase of a quality asset at a fair price.
\end{itemize}
Citations

\begin{itemize}
    \item https://investor.fm/unpacking-buffetts-investment-philosophy-a-personal-perspective/
    \item https://www.binance.com/en/square/post/18537661287098
    \item https://www.fool.com/investing/how-to-invest/famous-investors/charlie-munger/
    \item https://www.investinassets.net/p/10-lessons-from-charlie-munger
    \item https://www.ruleoneinvesting.com/blog/personal-development/5-charlie-munger-investing-lessons-every-value-investor-should-live-by
    \item https://www.gainify.io/blog/duan-yongping-portfolio
    \item https://www.iiflcapital.com/blog/personal-finance/lessons-from-warren-buffetts-investment-philosophy
    \item https://www.binance.com/en/square/post/13966098863626
    \item https://simplyethical.com/blog/warren-buffetts-investment-tenets/
    \item https://hbr.org/1996/01/what-i-learned-from-warren-buffett
    \item https://www.investopedia.com/articles/01/071801.asp
    \item https://www.alphaexponent.net/p/22-duan-the-dilettante
    \item https://finance.yahoo.com/news/charlie-mungers-timeless-investment-advice-113042652.html
    \item \url{https://link.springer.com/content/pdf/10.1007/978-981-95-0545-6_10.pdf}
\end{itemize}
\end{document}